\documentclass{article}

\PassOptionsToPackage{numbers, compress}{natbib}

\usepackage[preprint]{neurips_2020}

\usepackage[utf8]{inputenc} 
\usepackage[T1]{fontenc}    
\usepackage{hyperref}       
\usepackage{url}            
\usepackage{booktabs}       
\usepackage{amsfonts}       
\usepackage{nicefrac}       
\usepackage{microtype}      
\usepackage{graphicx}
\usepackage{amsmath}
\usepackage{comment}

\def\ie{\emph{i.e.}}
\def\eg{\emph{e.g.}}

\title{Learnable Sampling 3D Convolution for Video Enhancement and Action Recognition}

\author{%
  Shuyang Gu \\
  USTC \\
  \texttt{gsy777@mail.ustc.edu.cn}\\
  \And
  Jianmin Bao \\
  Microsoft Research \\
  \texttt{jianbao@microsof.com}  \\
  \And
  Dong Chen \\
  Microsoft Research \\
  \texttt{doch@microsoft.com}  \\
}

\begin{document}

\maketitle

\begin{abstract}
A key challenge in video enhancement and action recognition is to fuse useful information from neighboring frames. Recent works suggest establishing accurate correspondences between neighboring frames before fusing temporal information.
However, the generated results heavily depend on the quality of correspondence estimation.
In this paper, we propose a more robust solution: \emph{sampling and fusing multi-level features} across neighborhood frames to generate the results. Based on this idea, we introduce a new module to improve the capability of 3D convolution, namely, learnable sampling 3D convolution (\emph{LS3D-Conv}). We add learnable 2D offsets to 3D convolution which aims to sample locations on spatial feature maps across frames. The offsets can be learned for specific tasks. The \emph{LS3D-Conv} can flexibly replace 3D convolution layers in existing 3D networks and get new architectures, which learns the sampling at multiple feature levels. The experiments on video interpolation, video super-resolution, video denoising, and action recognition demonstrate the effectiveness of our approach.
\end{abstract}

\section{Introduction}

There is increasing interest in video interpolation~\cite{sullivan1991motion,zitnick2004high,jiang2018super,niklaus2017video,niklaus2018context}, video super-resolution~\cite{liu2011bayesian,shi2016real,kappeler2016video,sajjadi2018frame,li2019fast}  and video denoising~\cite{dugad1999video,dabov2007image,mahmoudi2005fast,ji2010robust}. The aim of these tasks is to recover a high quality video from an input video suffering from degradation (low-bit rate, low resolution or noise). The key to the success of video interpolation and restoration is to collect and fuse useful information from neighboring frames.

Most existing approaches adopt the \emph{alignment and fusing} strategies. Some optical flow based methods~\cite{ohnishi2018hierarchical,chen2017coherent,wang2018video,jiang2018super,xue2019video} align the reference and its neighboring frames by explicitly estimating the motion/flow vector for each pixel. However, estimating optical flow remains to be a challenging problem due to fast-moving objects, occlusions and motion blur. Besides, the flow estimation network is usually trained on synthetic datasets. Their generalization ability to the real world is limited. The other kinds of approaches to achieve implicit motion compensation are dynamic filtering~\cite{jo2018deep,liu2017robust} or deformable convolution~\cite{tian2018tdan,xu2019learning}. Without supervision, it is too difficult to estimate accurate motion. The inaccurate motion may cause incorrect fusion of pixels from neighboring frames, resulting in ghosting or blurring.

Recently, non-local neural networks~\cite{wang2018non} is applied for video super-resolution~\cite{yi2019progressive}. The non-local operation fuses all possible pixels by computing the correlations with neighboring frames. But it usually fuses patches with similar appearance, instead of the same semantic instance. The computation cost of non-local operation is also relatively heavy for capturing long-range dependencies.

\begin{figure}[t]
	\centering
	\includegraphics[width=0.8\columnwidth]{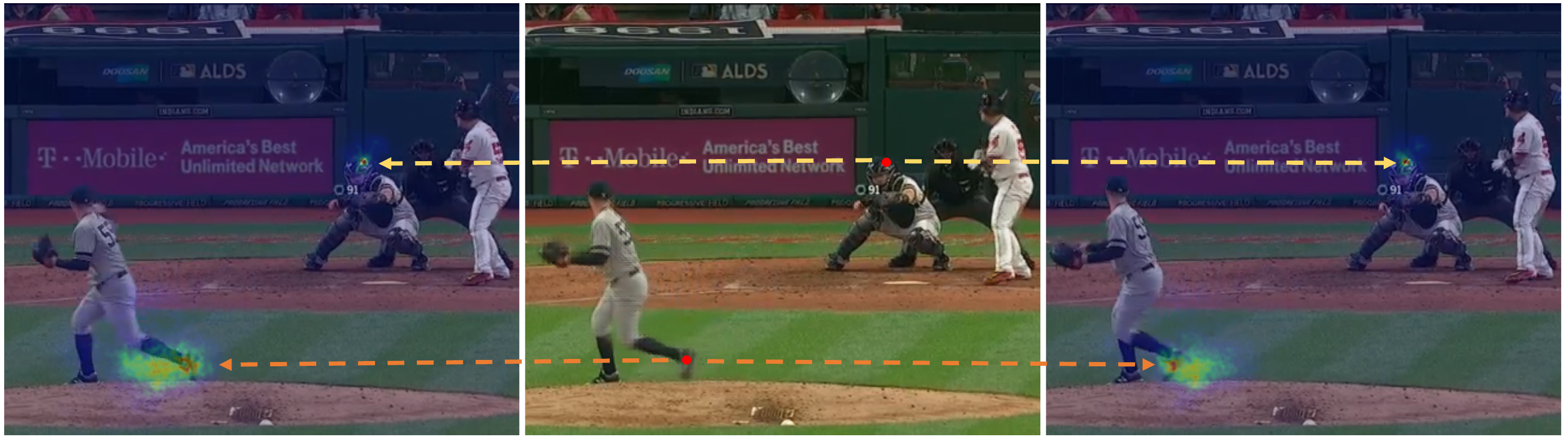}
	\caption{The medium is the interpolation result of two input frames(left and right), we show the sampling locations on the input frames correspond to each output pixel(red points) from our proposed \emph{LS3D-ConvNet}. We can observe that learnt sampling positions is more concentrated in regions with small motion but more scattered for areas with large motion.}
	\label{fig:intro}
\end{figure}

\begin{table}
	\label{tab:comparsion_of_different_methods}
	\small
	\centering
	\begin{tabular}{c|c|c|c}
		module & dimenstion & learned offset or regular grid & learned kernel or interpolation \\ \hline
		2D Convolution & 2D & regular grid & N/A \\
		3D Convolution & 3D & regular grid & N/A \\
		2D Deformable Conv & 2D & learned offset & bilinear \\
		SDC-Net/MEMC-Net & 2D & learned offset & learned kernel \\
		LS3D-Conv (ours) & 3D & learned offset & learned kernel + bilinear \\ \hline
	\end{tabular}
	\caption{Comparsion of different methods.}
	\vspace{-0.5cm}
\end{table}


In this paper, we propose a novel end-to-end deep neural network which is composed of a set of \emph{learnable sampling 3D convolution} (\emph{LS3D-Conv}) modules for this task. Our network learns to \emph{sample and fuse multi-level features} from the neighboring frames to generate the video, which is the key difference from the existing work. In every level, \emph{LS3D-Conv} learns how to sample valuable features from adjacent frames, and then automatically fuses the collected feature candidates using a 3D convolution. The aggregated features are used for sampling in the next level. By iterations, the network finally outputs a reconstructed frame.

Inspired by the parametrization in 2D deformable convolution~\cite{dai2017deformable}, the sampling in neighboring frames can be modeled by 2D offsets around each location in nearby frames. Furthermore, we add an importance scalar for each sampling location to indicate the importance of the feature from that location. At every level, the network will directly learn such a set of 2D offsets and an importance scalar for each location.

Compared with the traditional 3D convolution~\cite{ji20123d}, or flow-guided convolution~\cite{zhao2018trajectory}, \emph{LS3D-Conv} brings several advantages: (a) In general, flow-based approaches may fail to find accurate correspondences across frames especially for large motion, or motion blur. In this case, \emph{LS3D-Conv} only needs the collected samples to cover valuable (or best matching) samples, which relaxes the requirement of matching accuracy; (b) \emph{LS3D-Conv} can learn and update the sampling locations for the target tasks; (c) The sampling strategy learned by \emph{LS3D-Conv} can be adapted to the confidence of correspondence estimation, which is more robust than single-level deformable convolution framework~\cite{tian2018tdan,xu2019learning}. As the example shown in Figure~\ref{fig:intro}, where the learned sampling is more concentrated in regions with high-confidence motion estimation and seems to be more scattered for areas with large motion, or motion blur.

The experiments show the effectiveness of our proposed \emph{LS3D-Conv} in video interpolation, video super-resolution, and video denoising. Furthermore, the proposed \emph{LS3D-Conv} operator can also be flexibly applied to popular action recognition backbone to boost the performance.

%
%
%
%
%
%


\section{Related work}

We briefly summarize the most related works for video enhancement. In general, most of these works can be divided into three types. The first is to use optical flow~\cite{horn1981determining,barron1994performance} to estimate the correspondence between neighboring frames and generate the result. The second is to leverage some new techniques(\eg non-local~\cite{wang2018non}, 2D deformable convolution~\cite{dai2017deformable,zhu2017deep}) to build the correspondence and reconstruct the result. The third is to directly apply 3D CNNs~\cite{tran2015learning}. Next, we will talk about these works in detail.

\noindent \textbf{Flow-based methods.}
Recent advances in optical flow estimation methods like FlowNet~\cite{dosovitskiy2015flownet}, EpicFlow~\cite{revaud2015epicflow}, FlowNet2.0~\cite{ilg2017flownet}, and PWCNet~\cite{sun2018pwc} arise many optical-based methods for video enhancement and action recognition. For example, Deep Voxel Flow~\cite{liu2017video} proposes an end-to-end deep network to learn how to borrow voxels from nearby frames for video frame synthesis. For high frame rate video interpolation, Super SloMo~\cite{jiang2018super} proposes an end-to-end convolutional neural network for variable-length multi-frame video interpolation. More recent work TOFlow~\cite{xue2019video} proposes task-oriented flow for accurate motion estimation and the flow is learned from target tasks. Although the flow estimation can be learned from target tasks, it still suffers from the fast-moving object, occlusions, and motion blur.

\noindent \textbf{Build correspondence beyond flow.} Instead of relying on flow estimation to build the correspondence. Several works have incorporated a learned motion estimation network in burst processiong~\cite{tao2017detail,jo2018deep,gu2018arbitrary,gu2019mask}. The recently proposed non-local blocks are also applied to build the long-range correspondence for video super-resolution~\cite{yi2019progressive} or video denoising~\cite{davy2019non}. From another point of view, video interpolation can also be formulated as convolution operations. Following this idea, recent works AdaConv~\cite{niklaus2017adaptive} and SepConv~\cite{niklaus2017video} estimate spatially-adaptive convolutional kernels for each output pixel. Moreover, recent works~\cite{tian2018tdan,xu2019learning,wang2019deformable,wang2019edvr} adopt deformable convolution to align the features between two adjacent frames for video super-resolution. 

\noindent \textbf{3D CNN based methods} It is a natural idea to extend 2D CNN to 3D CNN for video-related tasks. VESPCN~\cite{caballero2017real} apply a spatio-temporal sub-pixel convolution networks for real-time video super-resolution. 3DSRNet~\cite{kim20183dsrnet} proposes a 3D CNN based framework for video super-resolution. More recent work FSTRN~\cite{li2019fast} proposes an efficient 3D convolution based operator for video super-resolution, which achieves impressive results on video super-resolution. We highlight the difference compared with these methods in Table~\ref{tab:comparsion_of_different_methods}.

We also notice that several approaches do not belong to these types. For example, DAIN~\cite{bao2019depth} uses extra information in videos like depth for video interpolation. IM-Net~\cite{peleg2019net} proposes an end-to-end framework for high resolution video frame interpolation and formulate interpolated motions estimation as a classification problem. DeepSR-ITM~\cite{kim2019deep} proposes a joint super-resolution and inverse tone-mapping framework which is able to restore fine details in videos.

\begin{figure*}[t]
	\centering
	\includegraphics[width=1.0\columnwidth]{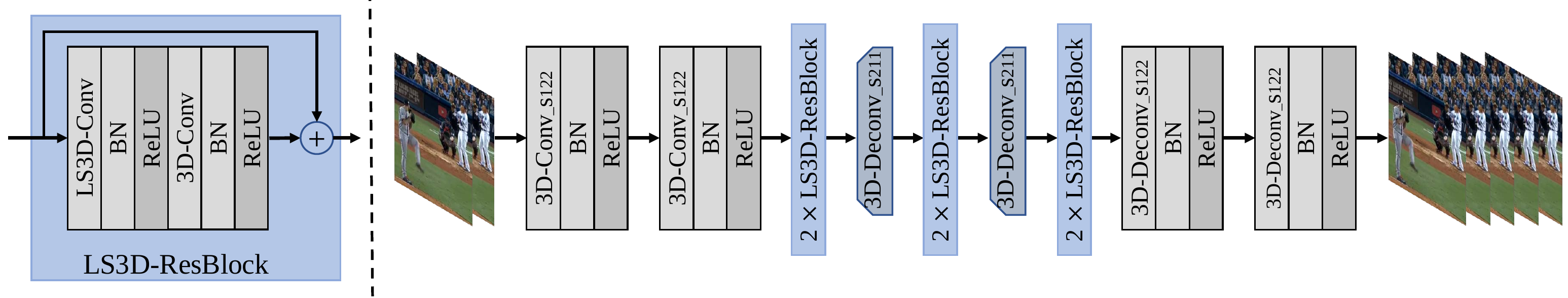}
	\caption{Illustration of our proposed LS3D-ResBlock and \emph{VI-LS3D-ConvNets} architectures for video interpolation, the S after convolution or deconvolution denotes the stride used at T, H, W dimension, respectively.}
	\label{fig:LS3DConv_Nets_for_video_interpolation}
	\vspace{-0.3cm}
\end{figure*}


\section{Methods}


In this paper, we propose a novel end-to-end framework composed of a set of \emph{learnable sampling 3D convolution} (\emph{LS3D-Conv}) modules for video enhancement and action recognition. The proposed framework aims to \emph{sample and fuse multi-level features} from neighboring frames for the target task. Therefore, we first take the video interpolation task as an example and introduce the proposed architecture. Then we introduce how our proposed LS3D-Conv learn to sample and fuse valuable feature for the next level. Finally, we discuss and clarify the relations and differences with some existing works.





\subsection{Learnable Sampling 3D Convolution Network}
Although 3D CNNs~\cite{ji20123d,qiu2017learning} are widely used in video-related tasks, we notice that few 3D CNNs architectures are proposed for video interpolation. For the integrity of the paper, we first detail the network architecture of the proposed method for video interpolation. As shown in Figure~\ref{fig:LS3DConv_Nets_for_video_interpolation}(b), our method mainly consists of three components: (1) An encoder like component consists of two 3D convolution layers (strides are $1, 2, 2$ for T, H, W, respectively). to reduce the spatial input size. (2) Interpolation component consists of $6$ learnable sampling 3D-ResBlocks, \ie,  \emph{LS3D-ResBlock}, which will be introduced later and two 3D deconvolution layers (strides are $2, 1, 1$ for T, H, W, respectively). Two 3D deconvolution layer are placed after the second and the forth LS3D-Resblocks, respectively. They are used for temporal upsampling. (3) A decoder like component uses two 3D deconvolution layers (strides are $1, 2, 2$ for T, H, W, respectively) to reconstruct the final results. The proposed  LS3D-Convolution network can interpolate $3$ in-between frames for two input frames. Our framework is simple and general. It can also be used for other video tasks, \eg, video super-resolution, video denoising, video recognition, and so on. We only need to change the third component and the temporal stride in 3D deconvolution layers as needed.


The proposed LS3D-ResBlock structure is shown in Figure~\ref{fig:LS3DConv_Nets_for_video_interpolation}(a). It applies a newly designed residual block, in which we replace the first convolution layer in the 3D-ResBlock with the learnable sampling 3D convolution (LS3D-Conv). These LS3D-Conv layer in the network will learn how to sample valuable features from adjacent frames,  and then automatically fuses the collected feature candidates at multi-level. We will introduce the technical details of how to learn to sample and fuse valuable features at each level in the following section.

\subsection{Learnable sampling 3D convolutions}

The main novelty of learnable sampling 3D convolutions relies on the learnable process of sampling and fusing of the input features of neighboring frames. We will describe these two steps in detail in the following part.

\noindent \textbf{Sampling} Let us first consider a conventional 3D convolution operator. Given input feature maps $\{{x}_t\}$, where ${x}_t \in \mathbb{R}^{C \times H \times W}$ is the feature of the $t$-th frame,  suppose the 3D convolution kernel $\mathbf{w} \in \mathbb{R}^{C \times 3 \times 3 \times 3}$. For a output spatial location $\mathbf{p}$, the sampling location for the traditional 3D convolution is $\mathbf{p} + \mathbf{p}^n$ on the input feature map across frame $t + \tau$. where $\mathbf{p}^n$ enumerates $3 \times 3$ spatial locations as a standard 2D convolution: $p^n \in \{(-1, -1), (-1, 0), \ldots ,(1, 1)\}$, and $\tau$ enumerates the temporal dimension $\tau \in \{-1, 0, 1\}$.

However, Sampling features at a fixed grid around location $\mathbf{p}$ across neighboring frames usually fail to provide valuable information for reconstruction in case of motions. To address this problem, we add additional 2D offsets $\Delta \mathbf{p}^n_{t+\tau}$ to the sampling locations $\mathbf{p}^n$ at frame $t+\tau$. Thus the sampling positions becomes: $\mathbf{p} + \mathbf{p}^n + \Delta \mathbf{p}^n_{t+\tau}$. 

In this case, the sampling locations on input feature maps at frame $t + \tau$ can be  irregular locations $\mathbf{p} + \mathbf{p}^n + \Delta\mathbf{p}_{t+\tau}^n$. And the offsets $\Delta\mathbf{p}_{t+\tau}^n$  can be obtained by directly adding a traditional 3D convolution layer on the input feature map. So the sampling locations can be learned by the input feature map.

Considering the offsets $\Delta\mathbf{p}_{t+\tau}^n$ is possibly becomes fractional, therefore, the corresponding feature $\mathbf{x}_{t+\tau}(\mathbf{\widetilde{p}})$, where $\mathbf{\widetilde{p}} = \mathbf{p} + \mathbf{p}^n + \Delta\mathbf{p}_{t+\tau}^n$,  is calculated by bilinear interpolation with a specific sampling kernel $G$:

\begin{equation}
\mathbf{x}_{t+\tau}(\mathbf{\widetilde{p}})=\sum_\mathbf{q} G(\mathbf{q},\mathbf{\widetilde{p}})\cdot \mathbf{x}_{t+\tau}(\mathbf{q}).
\label{eq.bilinear_interpolation}
\end{equation}
Here $\mathbf{q}$ enumerates all integral spatial locations in the feature map $\mathbf{x}_{t+\tau}$, and G is the bilinear interpolation kernel.

\noindent \textbf{Fusing} After sampling input features at the sampling positions $\mathbf{p} + \mathbf{p}^n + \Delta \mathbf{p}^n_{t+\tau}$. The next step for our method is to fuse these features. Here we use the 3D convolution operator for feature fusion. So the output feature $\mathbf{y}_t(\mathbf{p})$ can be calculated as:

\begin{equation}
\label{eqn:learned_sampling_conv_wo_importance}
\mathbf{y}_t(\mathbf{p}) = \sum\limits_{\tau = -1}^1 \sum\limits_{n = 1 }^9 \mathbf{w}_{\tau}(\mathbf{p}^n) \mathbf{x}_{t+\tau}(\mathbf{p} + \mathbf{p}^n + \Delta\mathbf{p}_{t+\tau}^n),
\end{equation}

where $\mathbf{w}_{\tau}$ is the weight of kernel for frame $t+\tau$. Besides, the sampling features may contribute differently to the final results. So we use an independent importance scalar $ \mathbf{m}^n_{t+\tau}$ in the range of $[0, 1]$ for each sampling locations across frames, then Eq.~\eqref{eqn:learned_sampling_conv_wo_importance} becomes

\begin{equation}
\label{eqn:learned_sampling_conv}
\mathbf{y}_t(\mathbf{p}) = \sum\limits_{\tau = -1}^1 \sum\limits_{n = 1 }^9 \mathbf{m}^n_{t+\tau} \mathbf{w}_{\tau}(\mathbf{p}^n) \mathbf{x}_{t+\tau}(\mathbf{p} + \mathbf{p}^n + \Delta\mathbf{p}_{t+\tau}^n) .
\end{equation}

The importance scalar can also be obtained by directly adding a traditional 3D convolution layer on the same input feature map. During the training stage, the offsets and the importance scalar are learned from the target tasks.

\subsection{In Context of Related Works}
In this section, we will detail the relations and difference of our proposed method with some existing works.

\noindent \textbf{Particle Filter~\cite{gustafsson2002particle}} Particle filter is a very popular technique for tracking tasks in videos. It basically goes through four basic steps: (1) initialization; (2) measurement update; (3) resampling and (4) prediction. Our proposed LS3D-Conv network shares similar steps in the inference stage: (1) At the first LS3D-Conv layer, we obtain an initial sampling location; (2) We fuse these sampled features and obtain an updated feature map; (3) Based on the new feature map, the next LS3D-Conv layer update the sampling location which is more suitable for the target; and (4) Repeat step (2)-(3) layer by layer till the end of the network. 

\noindent \textbf{Deformable Convolution~\cite{dai2017deformable,zhu2019deformable}} Deformable ConvNets augment the sampling locations in the 2D convolution with learnable offsets and modulations to handle geometric variations. Technically, our work can be viewed as extending the 2D deformable convolution to 3D, but still retaining the 2D offsets. But we have three differences compared to ~\cite{dai2017deformable,zhu2019deformable}. (1) The function of the offsets is different. Our LS3D-Conv aims to sampling meaningful features across neighboring frames, instead of handling geometric variations as~\cite{dai2017deformable,zhu2019deformable}. (2) Our offsets are not motion vectors. Comparing with~\cite{tian2018tdan,xu2019learning} that use deformable convolution to align features across frames, the offsets learned from a single LS3D-Conv layer seems meaningless. But when multiple LS3D-Conv layers work together, our method samples a large area to find useful features, as shown in Figure~\ref{fig:intro}. (3) To our knowledge, we are first to successfully use the deformable convolution for video interpolation tasks.

\noindent \textbf{TrajectoryNet~\cite{zhao2018trajectory}} This work applies estimated flow to 3D CNN for the motion features along the motion paths which can be aggregated. There are three main differences between this work and our method: (1) The sampling locations in different layers of trajectory convolution are the same, while our method uses different offsets for different layers. In Section~\ref{sec:understanding_LS3DConv-Nets}, we show the effectiveness of using multiple LS3D-Conv layers. (2) In trajectory convolution, the estimated trajectories are obtained by the other motion estimation network which introduces a huge additional computation cost. The approach used in our method to obtain offsets is much more efficient. (3) As mentioned above, the offsets learned by our method are not motion vectors.


\section{Experiments}
In this section, we first try to analyze the behavior of LS3D-Conv network in video interpolation tasks. Then, we show the results on video interpolation, video super-resolution, and video denoising tasks. Finally, we show that our proposed LS3D-Conv can even flexibly replace 3D-Conv in action recognition tasks to boost the performance.

\begin{figure*}[t]
	\centering
	\includegraphics[width=1.0\columnwidth]{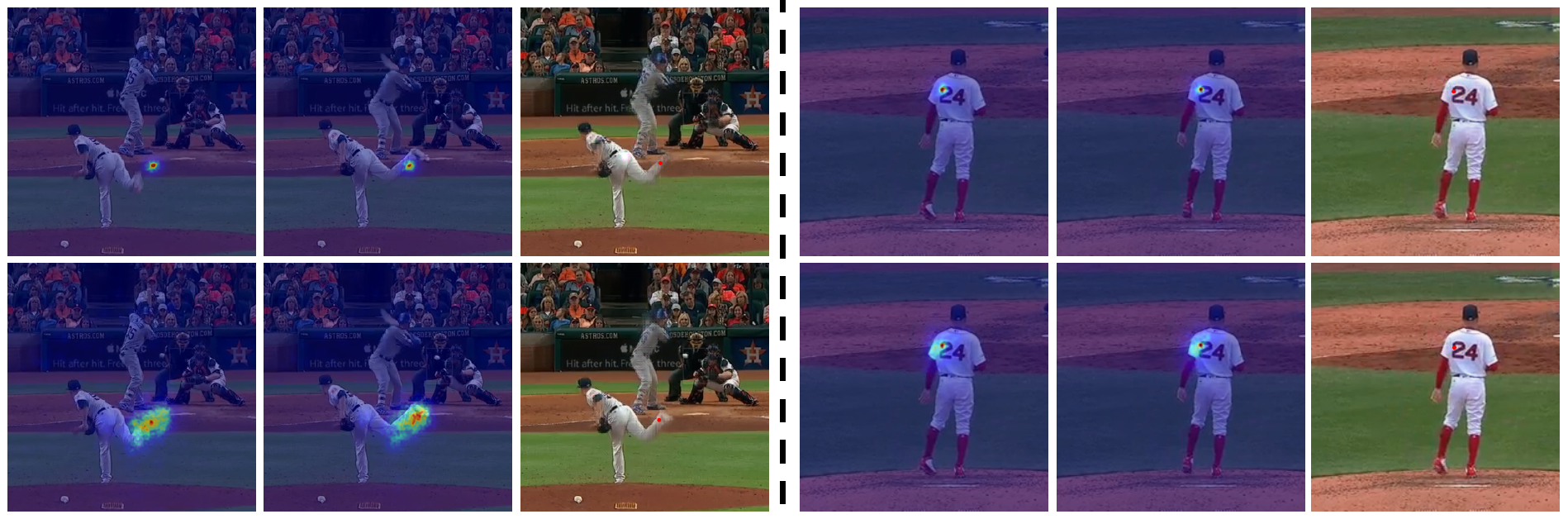}
	\caption{Each image triplet shows the two input frames and interpolation result of 3D-Conv network and LS3D-Conv network. We present the sampling locations on the input frames correspond to output pixel (red points) from the two models in large motion (left) and little motion (right).}
	\label{fig:abla1}
	\vspace{-0.3cm}
\end{figure*}


\subsection{Understanding LS3D-ConvNets}
\label{sec:understanding_LS3DConv-Nets}
To analyze and understand the behavior of the LS3D-Conv network, we adopt video interpolation task as an example to verify the learned sampling locations across frames. We train a baseline model with 3D-Conv network and compare it with the proposed LS3D-Conv network. All the experiments are conducted on MLB-Youtube datasets~\cite{mlbyoutube2018}. Please refer Section~\ref{sec:experiments_video_interpolation} for more implementation details.


\noindent \textbf{Visualization of the sampling locations.} To obtain the learned sampling locations for each model, we back-propagate from a chosen position on the interpolation output to get the gradient on the two input images. The gradient value can approximate the sampling frequency~\cite{luo2016understanding}. Figure~\ref{fig:abla1} illustrates several examples of the learned sampling locations of 3D-ConvNets and \emph{VI-LS3D-ConvNets}. We can observe that the sampling positions using the 3D-ConvNets are almost around the chosen positions. On the contrary, the learned sampling locations from LS3D-ConvNets can be adaptively adjusted according to different motions. When the output pixel is on an object with large motion, \eg, the leg in Figure~\ref{fig:abla1}, our method samples a large number of position candidates and fuses the valuable information to reconstruct the results. When the output pixel is on an object with small motion, our learnable sampling convolution samples a small number of position candidates around the output pixel.

\noindent \textbf{Which stage to use LS3D-Conv?}
We compare the results of adding LS3D-Conv into different stages. We apply LS3D-ResBlock in different layers and measure the video interpolation error. As shown in Table~\ref{table:results_stages_ablation}, LS3D-Conv at deep layer get better results. One possible explanation is that the feature from deeper layers contains stronger semantic information that helps to sample the correct position on the neighboring frames.


\noindent \textbf{Multi-level feature sampling leads to better results.}
Table~\ref{table:results_more_stages_ablation} shows the results of using LS3D-Conv in multi-level features. We train three models with different numbers of LS3D-Conv layers. Specifically, we use 2 LS3D-Conv layers(res5,6), 4 LS3D-Conv layers(res3,4,5,6), and all the 6 LS3D-Conv layers in the backbone. We can find that with more LS3D-Conv layers, the model can achieve better results.

\begin{table}[!t]
\begin{minipage}{\columnwidth}
	\begin{minipage}[t]{0.47\columnwidth}
		\small
		\centering
		\begin{tabular}{c|c|c}
			model & PSNR  &  SSIM \\
			\hline
			baseline & 29.35 & 0.942 \\
			\hline
			$\text{res1,2}$ & 29.70 & 0.943 \\
			\hline
			$\text{res3,4}$ & 30.45 & 0.951  \\
			\hline
			$\text{res5,6}$ & \textbf{30.97} & \textbf{0.954} \\
		\end{tabular}
	\makeatletter\def\@captype{table}\makeatother\caption{Comparison of results when adding learnable sampling 3D convolution into different stages.}
	\label{table:results_stages_ablation}
	\vspace{-0.5cm}
	\end{minipage}
    \quad \quad
	\begin{minipage}[t]{0.47\columnwidth}
		\small
		\centering
		\begin{tabular}{c|c|c}
			model & PSNR  &  SSIM \\
			\hline
			baseline & 29.35 & 0.942 \\
			\hline
			2-LS3D-Conv & 30.97 & 0.954 \\
			\hline
			4-LS3D-Conv & 31.70 & 0.962  \\
			\hline
			6-LS3D-Conv & \textbf{31.98} & \textbf{0.964} \\
		\end{tabular}
	\makeatletter\def\@captype{table}\makeatother\caption{Comparison of results when adding 2, 4,and 6 learnable sampling 3D convolution layers into model.}
	\label{table:results_more_stages_ablation}
	\vspace{-0.5cm}
	\end{minipage}
\end{minipage}
\end{table}

%



\begin{figure*}[t]
	\centering
	\includegraphics[width=1.0\columnwidth]{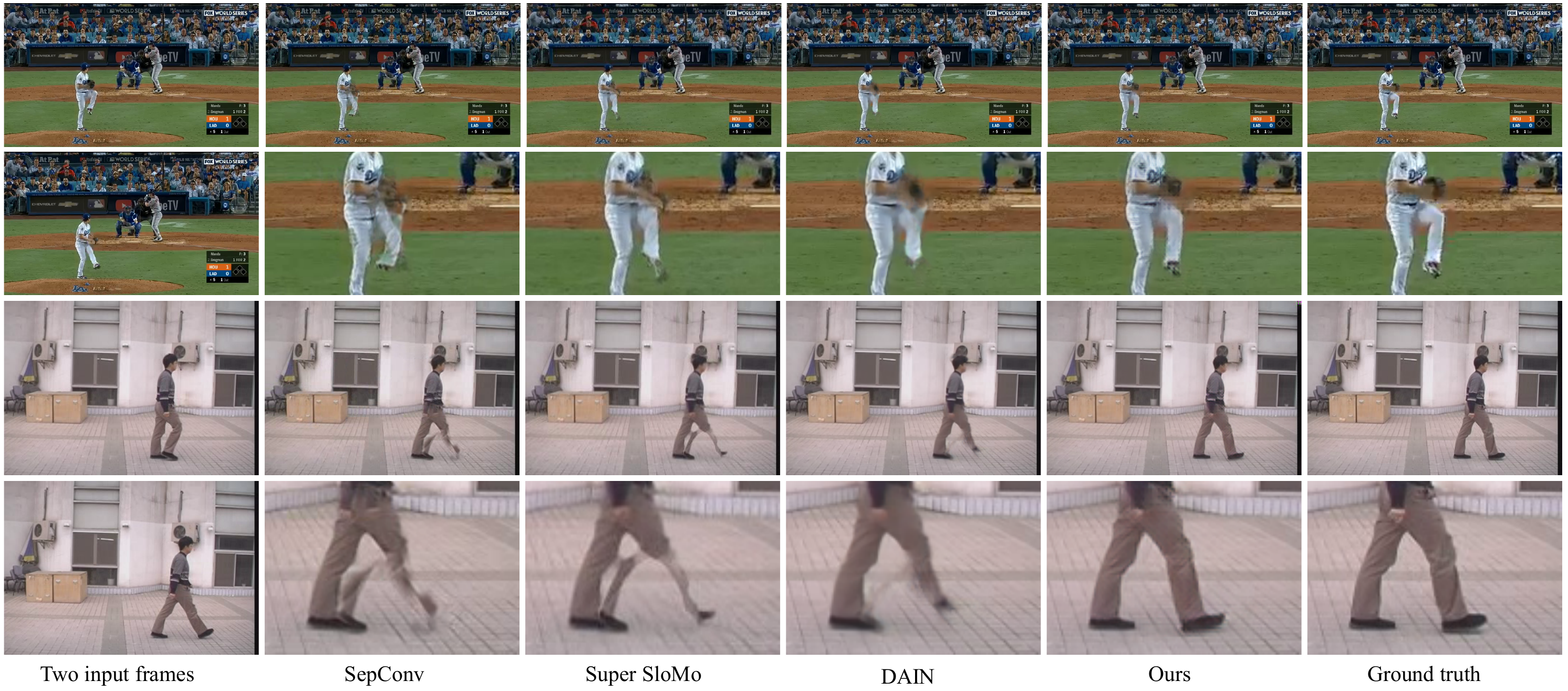}
	\caption{Qualitative video interpolation results comparison of SepConv~\cite{niklaus2017video}, Super SloMo~\cite{jiang2018super}, DAIN~\cite{bao2019depth} and our methods on MLB-Youtube datasets~\cite{mlbyoutube2018} and Gait Analysis dataset ~\cite{wang2003silhouette}.}
	\label{fig:inter_single_frame_results}
	\vspace{-0.6em}
\end{figure*}

\subsection{LS3D-ConvNets for Video Interpolation}
\label{sec:experiments_video_interpolation}

\begin{table}[t]
	\centering
	\small
	\begin{tabular}{c|c|c|c}
		Methods  & Cityscapes  & MLB-baseball & Gait  \\
		\hline
		SepConv~\cite{niklaus2017video}  & 27.32/0.819 & 28.87/0.926 & 33.01/0.947 \\
		\hline
		Super SloMo~\cite{jiang2018super} & 26.98/0.807 & 28.38/0.914 & 32.70/0.932 \\
		\hline
		DAIN~\cite{bao2019depth} & 27.48/0.830 & 30.31/0.954 & 35.02/0.964  \\
		\hline
		Ours & \textbf{27.62}/\textbf{0.837} & \textbf{31.98}/\textbf{0.964} & \textbf{35.82}/\textbf{0.978}  \\
	\end{tabular}
	\caption{Comparisons with state-of-the-art results on three datasets for video interpolation task, we report the PSNR and SSIM score of each method.}
	\label{table:inter_results}
	\vspace{-0.5em}
\end{table}


\begin{table*}[t]
	\centering
	\small
	\begin{tabular}{c|c|c|c|c|c|c}
		Methods  &  Dancing  & Treadmill  & Flag  & Fan & Turbine & Average \\
		\hline
		SRGAN~\cite{ledig2017photo} & 27.91 / 0.87  & 22.61 / 0.73  & 28.71 / 0.83  & 34.25 / 0.94 & 27.84 / 0.81 & 29.20 / 0.84 \\
		\hline
		BRCN~\cite{huang2015bidirectional} &  28.08 / 0.88 & 22.67 / 0.74 & 28.86 / 0.84 & 34.15 / 0.94 &27.63 / 0.82 & 29.16 / 0.85 \\
		\hline
		VESPCN~\cite{caballero2017real} &  27.89 / 0.86 & 22.46 / 0.74 & 29.01 / 0.85 & 34.40 / 0.94 & 28.19 / 0.83 & 29.40 / 0.85\\
		\hline
		FSTRN~\cite{li2019fast} & 28.66 / 0.89  & 23.06 / 0.76  & 29.81 / 0.88  & 34.79 / 0.95 & 28.57 / 0.84 & 29.95 / 0.87 \\
		\hline
		Ours & \textbf{29.06} / \textbf{0.91} &\textbf{23.13} / \textbf{0.77} & \textbf{29.98} / \textbf{0.89} & \textbf{35.07} / \textbf{0.95} & \textbf{28.82} / \textbf{0.85} & \textbf{30.17} / \textbf{0.88} \\
	\end{tabular}
	\caption{Comparison of the results on 25 YUV format benchmark, the performance is measured by PSNR/SSIM.} 
	\label{table:yuv-results}
\end{table*}

\begin{table}[!t]
	\begin{minipage}{\columnwidth}
		\begin{minipage}[t]{0.47\columnwidth}
			\small
			\centering
			\begin{tabular}{c|c|c}
				Methods  &  PSNR  & SSIM \\ 
				\hline
				Bicubic & 29.79  & 0.8483 \\
				\hline
				TOFlow~\cite{xue2019video} & 33.08  & 0.9054 \\
				\hline
				RCAN~\cite{zhang2018image} & 33.61  & 0.9101 \\
				\hline
				DUF~\cite{jo2018deep} & 34.33  & 0.9227 \\
				\hline
				3D-ConvNets & 33.25 & 0.9133 \\
				\hline
				Ours & \textbf{34.90} & \textbf{0.9295} \\
			\end{tabular}
			\makeatletter\def\@captype{table}\makeatother\caption{Quantitative results comparison with state-of-the-art methods of video super-resolution on Vimeo-90K dataset.}
			\label{table:vimeo-results}
			\vspace{-0.5cm}
		\end{minipage}
		\quad \quad
		\begin{minipage}[t]{0.47\columnwidth}
			\small
			\centering
			\begin{tabular}{c|c|c}
				Methods  &  Gaussian-15  & Gaussian-25 \\
				\hline
				Fixed Flow~\cite{xue2019video} & 36.25/0.9626  & 34.74/0.9411 \\
				\hline
				TOFlow~\cite{xue2019video} & 36.63/0.9628  & 34.89/0.9518 \\
				\hline
				3D-ConvNets & 36.35/0.9645 & 34.67/0.9452 \\
				\hline
				Ours & \textbf{36.90}/\textbf{0.9732}  & \textbf{35.23}/\textbf{0.9605} \\
			\end{tabular}
		    \vspace{0.2cm}
			\makeatletter\def\@captype{table}\makeatother\caption{Comparison of our approach with existing methods on video denoising task, the performance is measured by PSNR/SSIM.}
			\label{table:denoise-results}
			\vspace{-0.3cm}
		\end{minipage}
	\end{minipage}
\end{table}


\noindent \textbf{Training datasets.} We conduct video interpolation experiments on three datasets: Cityscapes datasets~\cite{cordts2016cityscapes}, MLB-Youtube datasets~\cite{mlbyoutube2018} and Gait Analysis dataset ~\cite{wang2003silhouette}. Cityscapes datasets have $2,974$ video clips of $17$-fps, each clip has 30 frames. MLB-Youtube dataset consists of $4290$ video clips of $30$-fps from $20$ baseball games, each clip contains diverse baseball activities such as swing, hit, ball, strike, and so on. And the gait datasets consists $240$ $30$-fps videos from 20 persons, each person has $12$ video sequences from different directions. The input and output resolutions for these three datasets are $256 \times 512$, $360 \times 720$ and $240 \times 352$ in our experiments, respectively. For all these datasets, we predict the in-between three frames by given two input frames.

\noindent \textbf{Training details.} For the training loss, we use $\mathcal{L}_1$ loss and two kinds of adversarial losses, a 2D adversarial loss, and a 3D adversarial loss to generate more realistic and coherent sequences. Please refer to supplementary material for more implementation details.


We illustrate the qualitative comparison results with the state-of-the-art method on MLB-Youtube and Gait Analysis datasets in Figure~\ref{fig:inter_single_frame_results}, in this figure, we only visualize the middle frame. We can find that our method has a better performance to handle large motion and occlusion. Table~\ref{table:inter_results} report the quantitative comparison results, our method achieves better performance compared with state-of-the-art methods. More video results are presented in the supplementary material.


\subsection{LS3D-ConvNets for Video Super-Resolution}

\noindent \textbf{Training datasets.} We conduct video super-resolution experiments on two datasets: 25 YUV format benchmark and Vimeo-90K~\cite{xue2019video}. 25 YUV format benchmark contains 25 YUV sequences for training and 5 sequences for test, which has been previously used in ~\cite{huang2015bidirectional,ledig2017photo,caballero2017real,li2019fast}. The Vimeo-90K dataset is a large, high-quality, and diverse dataset for video super-resolution, video denoising, and other video restoration tasks. The Vimeo-90K super-resolution benchmark consists of 91701 7-frame sequences with fixed resolution $448 \times 256$, extracted from 39K selected video clips from Vimeo-90K. For both datasets, we conduct our experiments with the upsampling scale of 4.

\noindent \textbf{Training details.} For the 25 YUV format benchmark, we follow the training and evaluation settings used in FSTRN~\cite{li2019fast}, and use a P3D based framework as backbone. We directly replace the $3 \times 1 \times 1$ 3D-Conv layer with LS3D-Conv. For the Vimeo-90K dataset~\cite{xue2019video}, we use a model based on image super-resolution model but with fewer ResBlocks, also the sub-pixel layer~\cite{shi2016real} is applied in the network. Please refer to the supplementary material for more details.

Table ~\ref{table:yuv-results} shows the quantitative comparison results on the 25 YUV format benchmark. We can find that our model achieves higher PSNR and SSIM score, which demonstrates the effectiveness of learnable sampling.




\subsection{LS3D-ConvNets for Video Denoising}

\noindent \textbf{Training datasets and details.}
We conduct the video denoising experiment on Vimeo-90K denoising benchmark~\cite{xue2019video}, which is the same as Vimeo-90K super-resolution benchmarks. We train and evaluate our method following TOFlow~\cite{xue2019video} with two kinds of noise, Gaussian noise with the standard deviation of 15 intensity levels(Gaussian-15) and Gaussian noise with the standard deviation of 25(Gaussian-25).


In Table~\ref{table:denoise-results}, We quantitatively compare the results of our method with Fixed Flow~\cite{xue2019video}, TOFlow~\cite{xue2019video}, and our baseline 3D-ConvNets. Our method achieve a higher PSNR and SSIM score. 

\begin{table}[t]
	\centering
	\footnotesize
	\begin{tabular}{c|c|c|c|c}
		Methods  &  Top1  & Top5 & params & GFLOPs \\
		\hline
		I3D & 72.0\%  & 89.9\% & 28.04M & 86.59 \\
		\hline
		I3D+LS3D-Conv & 72.9\%  & 90.6\% & 28.21M & 87.13 \\
		\hline
		I3D+non-local & 73.6\%  & 91.0\% & 35.40M & 100.10 \\
		\hline
		I3D+non-local+LS3D-Conv & \textbf{74.2\%}  & \textbf{91.2\%} & 35.57M & 100.64 \\
	\end{tabular}
	\caption{Action recognition results on Kinetics-400, reported on validation set. We use the ResNet-50 I3D model as our baseline.}
	\vspace{-0.5em}
	\label{tab:kenetics_results}
\end{table}


\subsection{LS3D-ConvNets for Action Recognition}

We also investigate our proposed LS3D-ConvNets for action recognition task on Kinetics-400~\cite{kay2017kinetics} dataset. Kinetics-400 contains $\sim$\noindent$246$k 
training videos and $20$k validation videos for $400$ human action categories. We train all models on the training set and evaluate on the validation set following ~\cite{carreira2017quo}. 
We choose two basic models for our baseline, ResNet-50 I3D~\cite{carreira2017quo} and ResNet-50 I3D model with non-local blocks~\cite{wang2018non}. We replace the $3 \times 1 \times 1$ 3D convolution in all bottleneck structure with our proposed learnable sampling 3D convolution in two baseline models. We train our models using the weights from the baseline model as initialization. The training procedure contains 45 epochs with an initial learning rate $0.001$ and it is divided by 10 for every $15$ epochs, the input size is $32 \times 224 \times 224$.

Table~\ref{tab:kenetics_results} shows the results of baseline models and the models with the proposed LS3D-Conv. The computational cost(GFLOPs) and the amount of learnable parameters are also reported. We can notice that our model (+LS3D-Conv) gets $0.9\%$ top-$1$ accuracy improvement over the simple baseline with negligible additional overhead. Although adding non-local blocks (+non-local) gets a better improvement over the baseline but comes with huge computation cost and model size. Our model is also compatible with non-local, the combination (+non-local+LS3D-Conv) can even get $0.6\%$ top-$1$ accuracy improvement over the model with non-local blocks (+non-local), which further shows the effectiveness of our proposed LS3D-ConvNets.



\section{Conclusion}
In this paper, we try to solve the video enhancement and action recognition in a \emph{sampling and fusing multi-level features} fashion. To achieve this, we propose a new operator called LS3D-Conv with learnable 2D offsets and importance scalar. The sampling locations and the importance scalar can be learned from target tasks. We can flexibly replace 3D convolution with our proposed LS3D-Conv in various tasks.
The experiments show the superiority of LS3D-Conv in video enhancement and action recognition. The proposed new operator can even be applied to popular action recognition frameworks to boost the performance.

{\small
	\bibliographystyle{ieee_fullname}
	\bibliography{egbib}
}

\end{document}